%% file: main.tex
\documentclass[10pt,twocolumn,letterpaper]{article}

\usepackage[pagenumbers]{wacv} 

\usepackage{graphicx}
\usepackage{amsmath}
\usepackage{amssymb}
\usepackage{booktabs}

\usepackage{multirow}
\usepackage{adjustbox}
\usepackage[table]{xcolor} 

\definecolor{1}{HTML}{5abc8c}
\definecolor{2}{HTML}{91d2b2}
\definecolor{3}{HTML}{d2ede0}

\usepackage[pagebackref,breaklinks,colorlinks]{hyperref}

\usepackage[capitalize]{cleveref}
\crefname{section}{Sec.}{Secs.}
\Crefname{section}{Section}{Sections}
\Crefname{table}{Table}{Tables}
\crefname{table}{Tab.}{Tabs.}

\newcommand{\NAME}{\textbf{SALVE}}

\def\wacvPaperID{1248} 
\def\confName{WACV}
\def\confYear{2025}

\begin{document}

\title{SALVE: A 3D Reconstruction Benchmark of Wounds \\from Consumer-grade Videos}

\author{Remi Chierchia\textsuperscript{1,2}, 
Leo Lebrat\textsuperscript{1,2}, 
David Ahmedt-Aristizabal\textsuperscript{1,2}, 
Olivier Salvado\textsuperscript{1}, \\
Clinton Fookes\textsuperscript{1}, 
Rodrigo Santa Cruz\textsuperscript{1,2} \\
Queensland University of Technology\textsuperscript{1}, 
CSIRO Data61\textsuperscript{2} \\
{\tt\small remi.chierchia@hdr.qut.edu.au}\\
}

\twocolumn[{
\maketitle
\begin{center}
    \captionsetup{type=figure}
    \includegraphics[width=\linewidth]{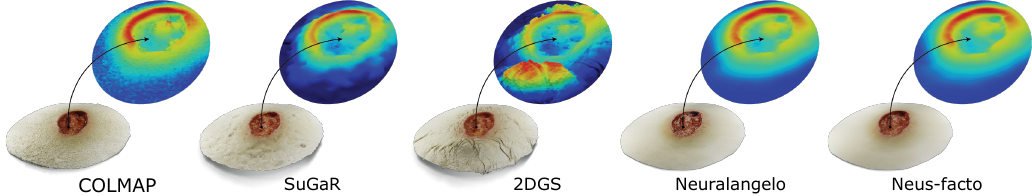}
    \captionof{figure}{Qualitative evaluation of six 3D surface reconstruction methods using our \NAME  \ dataset. On the left, we present COLMAP, a classic photogrammetric pipeline, which represents typical methods utilised for 3D reconstruction. 
    Next, we display two Gaussian splatting approaches SuGaR~\cite{guédon2023sugarsurfacealignedgaussiansplatting} and 2DGS~\cite{huang20242DGS}, followed by two NeRF approaches Neuralangelo~\cite{li2023neuralangelohighfidelityneuralsurface} and Neus-facto~\cite{Yu2022SDFStudio}. To highlight surface regularity\textemdash  important for wound analysis\textemdash we present depth color-coded from blue (far regions) to red (closer regions). }
\end{center}
}]

\begin{abstract}
Managing chronic wounds is a global challenge that can be alleviated by the adoption of automatic systems for clinical wound assessment from consumer-grade videos. While 2D image analysis approaches are insufficient for handling the 3D features of wounds, existing approaches utilizing 3D reconstruction methods have not been thoroughly evaluated. To address this gap, this paper presents a comprehensive study on 3D wound reconstruction from consumer-grade videos. Specifically, we introduce the \NAME~dataset, comprising video recordings of realistic wound phantoms captured with different cameras. Using this dataset, we assess the accuracy and precision of state-of-the-art methods for 3D reconstruction, ranging from traditional photogrammetry pipelines to advanced neural rendering approaches. In our experiments, we observe that photogrammetry approaches do not provide smooth surfaces suitable for precise clinical measurements of wounds. Neural rendering approaches show promise in addressing this issue, advancing the use of this technology in wound care practices.
We encourage the readers to visit the project page: \href{https://remichierchia.github.io/SALVE/}{SALVE}.
\end{abstract}

\section{Introduction}
\label{sec:intro}

Chronic wounds represent a significant health and economic burden worldwide~\cite{pacella2018solutions,wilkie2023determining}. Effective wound treatments depend on multiple wound clinical measurements, typically performed manually by specialized healthcare professionals~\cite{keast2004contents,nichols2015wound}. For instance, wound surface area is typically measured by performing planimetry of the wound bed~\cite{REIFS2023101185}. These procedures are not only invasive and cause patient discomfort but are also prone to errors due to ambiguous definitions of metrics and variations in professionals' skill levels. Moreover, this method is neither cost-effective nor scalable, as it requires specialized personnel who are often not available in remote areas.

\begin{figure}
    \centering
    \includegraphics[width=0.8\linewidth]{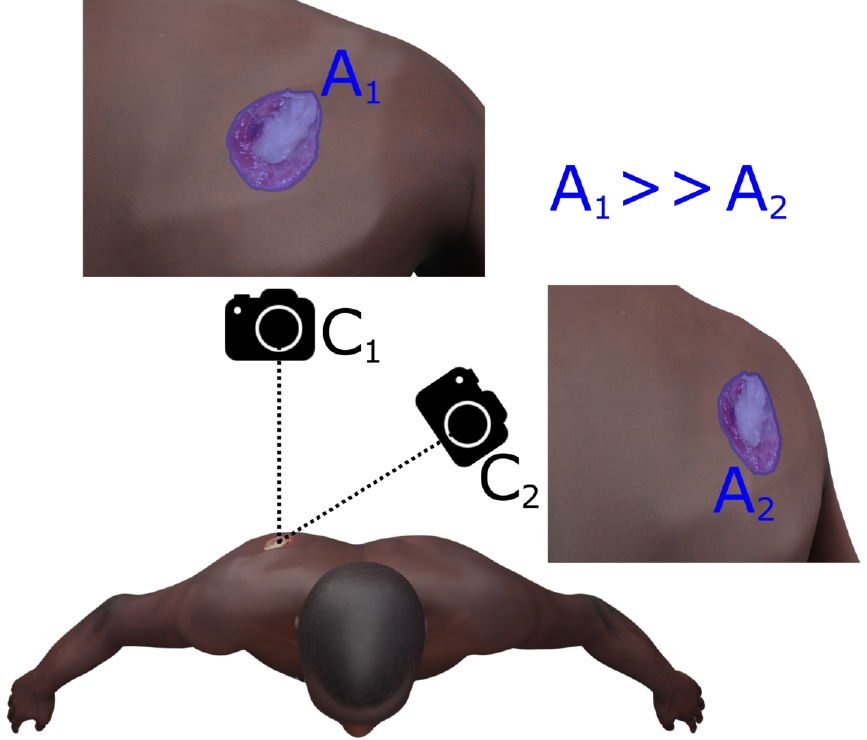}
    \caption{Limitation of 2D Wound Analysis: Images of a wound taken from different view perspectives will present significantly different wound areas.}
    \label{fig:2d3d}
\vspace{-6pt}
\end{figure}

Automatic measurement systems using computer vision and off-the-shelf cameras offer a promising avenue for wound care~\cite{anisuzzaman2022image,guarro2021wounds}. However, most existing commercial approaches~\cite{smartheal,imito} compute wound measurements solely from 2D images, which are inherently inaccurate for wounds with complex geometry or located on highly curved body parts such as the heel, toe, and lower leg. Also, 2D wound measurements are perspective-dependent. As illustrated in Figure~\ref{fig:2d3d}, wound measurements such as surface area can vary significantly when computed from images taken from different view angles. Furthermore, these approaches can not estimate wound depth, potentially overlooking a crucial aspect of the wound healing process.
%
%

In response to these limitations, 3D reconstruction techniques can be leveraged to generate 3D models from short videos depicting the wound from multiple viewpoints. Subsequently, measurements can be directly computed in 3D. This approach ensures that measurements conducted on the reconstructed geometry are independent of the view direction. Additionally, 3D analysis of wounds allows for the computation of richer wound biomarkers, which could streamline wound documentation~\cite{juszczyk2020wound}. Leveraging these prospects, several works have advocated for 3D wound documentation~\cite{juszczyk2020wound,Filko2018WoundCamera,albouy2006accurate,Juszczyk2019EvaluationWounds,lebrat2023syn3dwound,Barbosa2020AccurateMotion,Kumar2019AWounds,Liu2019WoundImages,Shirley2019AWounds}. However, these studies only considered previous generation 3D reconstruction frameworks, which have recently been surpassed by highly optimized photogrammetric toolboxes and recent neural rendering alternatives.

More importantly, no studies to date have thoroughly evaluated and compared the effectiveness of 3D reconstruction methods specifically applied to wound reconstruction from videos. To address this gap, we introduce a new dataset \NAME, designed to capture common challenges encountered in clinical settings, such as complex lighting conditions and varied image quality and resolution of realistic silicone wounds.
Using this dataset, we evaluate robust photogrammetry pipelines and modern neural rendering approaches for 3D reconstruction.
Our work includes a rigorous evaluation protocol that defines metrics and procedures to assess the geometric accuracy and precision of the evaluated reconstruction algorithms. 

We believe that this study is crucial to the development of reproducible research in the area of wound analysis and is a valuable contribution to the field of computer vision through the introduction of a new challenging task. 
Our benchmark highlights novel avenues for the use of neural 3D reconstruction in a medical setting and exhibits some limitations of current state-of-the-art (SOTA) methods when applied in this difficult scenario.
We highlight the robustness of different 3D reconstruction algorithms with respect to the quality of the acquisition device and demonstrate that some approaches show significant degradation in performance on lower-end devices.
We aim for this research to pave the way towards the development of affordable wound telehealth solutions and further reduce geographic inequalities in wound care access.

\section{Related Work}\label{sec:RelatedWork}

3D wound analysis has been a topic of interest for at least fifteen years~\cite{Treuillet2009Three-dimensionalCamera}, and its advantages over 2D methods and manual procedures have been highlighted in multiple clinical studies in wound care~\cite{Sirazitdinova2017SystemDevices,Kumar2019AWounds}. However, to the best of our knowledge, no study proposing a rigorous and unified evaluation of the accuracy and repeatability of state-of-the-art methods for 3D reconstruction applied to wound reconstruction from consumer-grade cameras exists.
Most existing works in this field focus on the repeatability and accuracy of specific wound measurements computed from images, pairs of images, or videos~\cite{Zenteno2018VolumeScanners,MirzaalianDastjerdi2019MeasuringAlgorithms,Sanchez-Jimenez2022SfM-3DULC:Area,Liu2019WoundImages,Barbosa2020AccurateMotion}. These studies typically use a single device and follow a particular acquisition protocol, without evaluating the 3D reconstruction itself. For instance, \cite{Zenteno2018VolumeScanners} conducted an inter-laboratory comparative study between photogrammetric and 3D scanner-based volume estimation of skin ulcers. Similarly, \cite{Liu2019WoundImages,MirzaalianDastjerdi2019MeasuringAlgorithms} performed experiments comparing wound area measurements using 2D and 3D methods. Lastly, \cite{Sanchez-Jimenez2022SfM-3DULC:Area} extended this analysis to nine different wound measurements performed by dermatologists and non-experts, using automatic and semi-automatic tools for wound measurements. The findings of these studies are limited to those specific measurements and cannot be generalized (e.g. different wound shapes, different devices, other wound etiology, and acquisition protocols,...).

Similar to our work, \cite{Treuillet2009Three-dimensionalCamera,Casas2015Low-costWounds,Shirley2019AWounds} evaluated the accuracy and repeatability of 3D reconstruction methods for wound analysis by comparing 3D models produced by these techniques with 3D models obtained with specialized 3D scanners. However, \cite{Treuillet2009Three-dimensionalCamera,Shirley2019AWounds} used a single artificial phantom created by a special effects artist as their input dataset. \cite{Casas2015Low-costWounds} used short video recordings of cutaneous leishmaniasis skin lesions, which are flat-looking and lack significant 3D features such as depth and complex shapes. These works also focused solely on a single acquisition device and used very simple photogrammetric pipelines. 
In contrast, our evaluation includes three types of wounds with rich 3D information for proper wound documentation, two acquisition devices, and state-of-the-art photogrammetric pipelines, as well as modern neural rendering approaches for 3D reconstruction. Additionally, unlike existing works, we will make our data and code publicly available for research purposes.

\begin{figure*}
    \centering
    \includegraphics[width=0.98\textwidth]{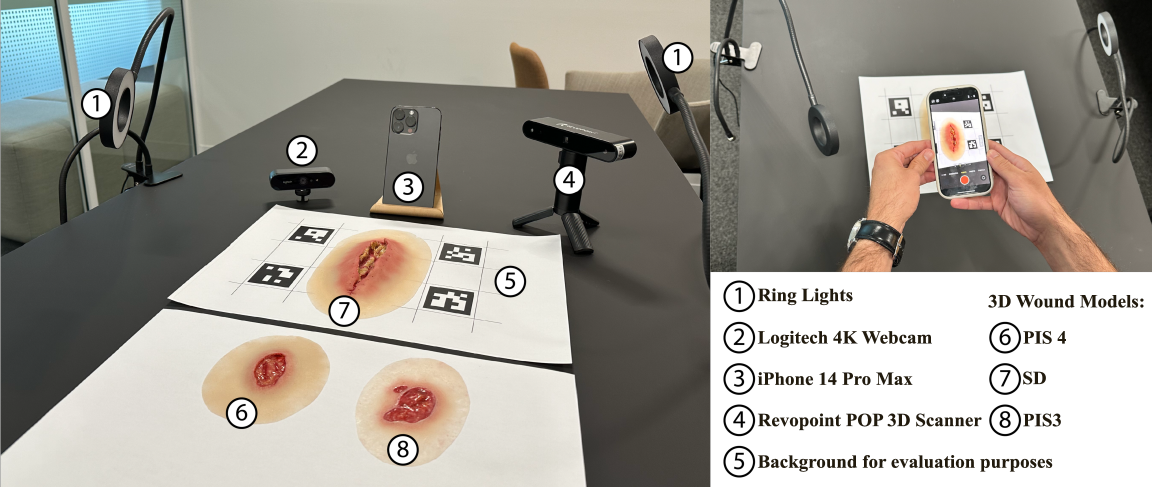}
    \caption{Materials used for acquiring the proposed dataset, including a picture of a recording session.}
    \label{fig:setup}
\vspace{-6pt}
\end{figure*}

Recently, studies have proposed evaluating 3D wound analysis methods in fully controlled synthetic 3D environments created with 3D computer graphics software or game engines. Lebrat et al. released Syn3DWound~\cite{lebrat2023syn3dwound} and Sinha et al. released DermSynth3D~\cite{SINHA2024103145}. 
Although one could utilize it to simulate some noise characteristics present in realistic settings, modeling the entire complexity of clinical environments paired with consumer device limitations is impractical.
This could lead to a reality gap in performance in the target application.


\section{Materials and Method}

\subsection{\NAME}
\label{sec:dataset}
To unify the evaluation protocol for 3D wound reconstruction and address the gap between synthetic and real-world acquisition, we propose \NAME, a 3D wound reconstruction benchmark from consumer-grade camera footage.
Given the ethical considerations associated with human data, and the time required to evaluate different data acquisition devices and settings, we opted for the use of high-quality silicone phantoms provided by TraumaSIM~\footnote{https://traumasim.com.au/ (a private company specialized in medically accurate simulations of realistic wounds to assist the training of healthcare professionals)}. 
In Figure~\ref{fig:setup}, we display the different wound types utilized in our study, denoted as SD (\textbf{S}urgical Wound \textbf{D}ehiscence), PIS3 (\textbf{P}ressure \textbf{I}njury \textbf{S}tage \textbf{3}), and PIS4 (\textbf{P}ressure \textbf{I}njury \textbf{S}tage \textbf{4}).

To encompass the variability of the acquisition device, we captured videos using both a smartphone (iPhone~14 Pro Max, 4K frames 2160 $\times$ 3840) and a webcam (Logitech 4K Webcam, 1K frames 1080 $\times$ 1920) under realistic lighting conditions, capturing reflections, shadows, and blurriness.
Accurate 3D ground-truth point clouds were obtained using the Revopoint POP 3D scanner~\footnote{https://global.revopoint3d.com/pages/face-3d-scanner-pop2}. 
Figure~\ref{fig:setup} shows an example of our acquisition materials and setup.

To mitigate noise and blurriness from video recordings, we devised a method that extracts a collection of sharp frames from each recorded video. We initially sample frames at 60 frames per second from each video. Then, we divide these frames into equal-sized consecutive frame sequences. From each sequence, we select the sharpest frame based on the variance of a 3x3 Laplacian filter applied to it. Using the recorded videos and this frame selection procedure, we generate various image datasets to support the evaluations outlined in Section~\ref{sec:exps_desc}.


\subsection{3D Reconstruction Methods }
\label{sec:3d_recon_methods}

The goal of 3D reconstruction is to estimate the 3D representation of a scene depicted in a set of images. Traditionally, this task is tackled by photogrammetry pipelines, which can be summarized in two main steps. 
The first step consists of solving a Structure from Motion (SfM)~\cite{snavely2006photo,agarwal2011building} problem, which produces a \textit{sparse} reconstruction. By matching features across images, an SfM algorithm regresses both the 3D structures and the camera matrices. 
The second step involves a Multi-View Stereo (MVS)~\cite{barnes2009patchmatch,moulon2016openmvg} approach, which establishes denser correspondences between image pairs using stereo matching and patch warping~\cite{furukawa2015multi}. 
Finally, a triangle mesh is generated by applying a meshing algorithm~\cite{kazhdan2013screened,levy2015geogram} to the obtained dense reconstruction.

As mentioned earlier, existing works on 3D reconstruction of wounds from videos follow this approach. For instance, \cite{Shirley2019AWounds} and \cite{Barbosa2020AccurateMotion} leverage well-established frameworks, such as COLMAP~\cite{schoenberger2016sfm,schoenberger2016mvs} and OpenSFM~\cite{opensfm}, respectively. However, they rely on accurate feature correspondences between images, which can be challenging in videos recorded in clinical settings due to complex lighting, moving shadows, and self-occlusions. These factors often lead to errors and discontinuities in the reconstructed 3D model.


A novel trend in 3D reconstruction involves using learnable implicit primitives paired~\cite{mildenhall2020nerf} with volumetric rendering~\cite{kajiya1984ray}. 
As detailed in \cite{mildenhall2020nerf}, these approaches express the color of a pixel $\hat{C}$ as,
\begin{equation}\label{eq:rendering}
\hat{C}(\mathbf{r}) \! = \!\! \int_{t_n}^{t_f} \! T(t)\sigma(\mathbf{r}(t))c(\mathbf{r}(t), \mathbf{d}) \text{d}t,  
\end{equation}
where $\mathbf{r}(t)$ is a time-parametrized ray for the pixel $\hat{C}$, $c$ is a learnable function depending on the 3D space location and the view direction $\mathbf{d}$, responsible for the color information of the neural radiance field, and $\sigma$ depends only on the spatial coordinate. 
In this formulation, the density $\sigma(\mathbf{x})$ can be interpreted as the probability of a ray being stopped at $\mathbf{x}$, and a 3D model can be obtained via iso-surface extraction algorithms. However, $\sigma$ is often implemented as an over-parameterized neural network without surface regularity constraints, which frequently results in 3D reconstructions with floaters or non-smooth surfaces~\cite{warburg2023nerfbusters}. These surface artifacts could compromise the reliability of clinical wound measurements, making them unsuitable for longitudinal tracking.

To improve the quality of the reconstructed geometry, multiple works~\cite{wang2021neus,wang2023neus2} propose deriving the density function $\sigma$ from a learnable Signed Distance Field (SDF), enabling surface regularity modelling within the standard volumetric rendering framework. For instance, in \cite{wang2021neus}, $\sigma$ is defined to be maximal at the zero-crossings of the SDF $f_{\theta}$,
\begin{align}\label{eqn:SDF2sigma}
 \sigma(\mathbf{x}) =& \max \left( \frac{-H'(f_{\theta}(\mathbf{x}))}{H(f_{\theta}(\mathbf{x}))} , 0\right),
\end{align}
where $H$ is the cumulative distribution function of the logistic distribution, and $\theta$ represents the model parameters to be learned through volumetric rendering. In this formulation, $f_\theta$ is subject to the eikonal constraint, and further regularizations, such as curvature~\cite{li2023neuralangelohighfidelityneuralsurface}, can be defined on this implicit representation of objects.
The training time for such approaches is in the order of several hours, which limits their applicability in some clinical settings. To reduce the training time associated with the discretization of the rendering Equation~\eqref{eq:rendering}, a novel family of methods has emerged. They propose an unstructured representation of radiance fields. Gaussian Splatting directly optimizes the position and variance of 3D Gaussians~\cite{kerbl20233dgaussiansplattingrealtime}. This representation remains volumetric but can be rasterized more efficiently by applying 2D projection and alpha-compositing.


However, Gaussian splatting does not allow for precise surface extraction. SuGaR~\cite{guédon2023sugarsurfacealignedgaussiansplatting} and 2DGS~\cite{huang20242DGS} proposed two different formulations amenable to surface extraction from a 3D sparse representation. Although similar in concept, they differ in their problem approach. 
SuGaR enforces additional constraints on the learned Gaussians to more accurately represent the geometry of the scene. More specifically, the Gaussians should have limited overlap with their neighbors, be as opaque as possible, and be as thin as possible in the normal direction of the surface. Under those assumptions, the neural density of the field can be approximated using,
\begin{equation}
    \hat{\sigma}(\mathbf{x}) = \exp \left( -\frac{1}{2 s^2_{g^\star}} \langle \mathbf{x} - \mathbf{\mu}_{g^\star}, \mathbf{d}_{g^\star} \rangle^2  \right),
\end{equation}
where ${g^\star}$ is the index of the closest Gaussian to $\mathbf{x}$, $s_{g^\star}$ is its smallest scaling factor, and  $\mathbf{d}_{g^\star}$ is the associated direction. By penalizing the 3D Gaussians to achieve such density, one can significantly enhance the geometric information. Finally, by deriving a set of points and their normals from the Gaussian density, one can employ Poisson reconstruction to swiftly extract a surface mesh with improved quality.

On the other hand, 2DGS proposes to remove the extra dimension $\mathbf{d}_{g^\star}$ and replace the 3D Gaussian with a 2D Gaussian, a multiview-consistent primitive more suited for surface extraction. Both of these explicit formulations reduce runtimes compared to their continuous SDF counterpart. However, they may fall short of providing the regularity required for our application.

\paragraph{\textbf{Benchmarked methods:}} 
We compare three proven photogrammetry pipelines: COLMAP~\cite{COLMAP} (CM), COLMAP equipped with LightGlue feature matching~\cite{lindenberger2023lightglue} (LGCM), and Meshroom~\cite{alicevision2021} (MR).
Additionally, leveraging the computed image poses from LGCM, we benchmark recent AI-based methods for 3D reconstruction. 
These include Neural Signed Field techniques like NeusFacto~\cite{Yu2022SDFStudio} and NeuralAngelo~\cite{li2023neuralangelohighfidelityneuralsurface}, as well as discrete explicit approaches such as 3D Gaussian Splatting (3DGS)~\cite{kerbl20233dgaussiansplattingrealtime} for rendering and SuGaR\cite{guédon2023sugarsurfacealignedgaussiansplatting} or 2D Gaussian Splatting~\cite{huang20242DGS} for 3D reconstruction.
Finally, we experimented with fast methods using Neural Hash encoding such as InstantNGP~\cite{muller2022instant} for rendering and NeuS2~\cite{wang2023neus2}, its SDF-based variant. We also explored SfM-free approaches like DUSt3R~\cite{wang2023dust3rgeometric3dvision}, an end-to-end stereo 3D reconstruction method.

%

%

\input{tables/accuracy_plus_resources}

\subsection{Experiments Description}
\label{sec:exps_desc}

For 3D wound reconstruction methods to be valuable in research and clinical settings, they must be both accurate and precise. Accuracy assesses how closely a reconstruction matches the ground truth, while precision (or reliability and repeatability) measures the consistency between repeated reconstructions. This consistency is essential for verifying the robustness to varying input data and ensuring suitability for tracking wound measurements over time. Additionally, these methods can serve as powerful tools for wound visualization. We evaluate the approaches above based on these three criteria, which are detailed below.

\paragraph{Accuracy Evaluation:} From each recorded video in our dataset, we initially curated a set of fifty frames using the frame selection procedure detailed in Section~\ref{sec:dataset}. We chose fifty frames as we observed that adding more frames did not enhance wound visibility and sometimes compromised the accuracy of SfM pose estimation due to the inclusion of blurry frames. Using these frames, we generated 3D models of the wounds using the methods outlined in Section~\ref{sec:3d_recon_methods} and compared their results to the ground truth obtained with the specialized 3D scanner detailed in Section~\ref{sec:dataset}. The outcomes of this experiment are presented in Table~\ref{tab:acc_resources}.

\paragraph{Precision Evaluation:} We evaluate the consistency of 3D reconstruction methods across different camera devices (inter-device) and recording attempts (inter-recording). From each video in our dataset, we selected 150 sharp frames using the procedure outlined in Section~\ref{sec:dataset}. These frames were then temporally divided into fifty sub-sequences, each containing three frames. Within each sub-sequence, the frames were randomly assigned to separate splits, resulting in three data splits of fifty frames for each video.

For inter-device precision, we compare all pairs of 3D reconstructions generated by each method using data splits generated from different recording devices for the same wound. To assess inter-recording precision, we focus only on data splits derived from videos recorded with the iPhone. Then, we compare all pairs of 3D reconstructions produced by each method from these data splits of the same wound. The results of this experiment are presented in Table~\ref{tab:prec_new}.

\input{tables/precision_new}

\paragraph{Rendering Quality:} To evaluate the suitability of the 3D reconstruction methods as visualization tools (excluding the photogrammetric pipelines), we assessed their rendering capability for view perspectives not present in the input training images. To this end, we extract 75 frames from each video in our dataset using our frame selection procedure, divide them into 50 training images and 25 test images with different poses and visual content, as suggested in \cite{Xiao:CVPR24:NeRFDirector}. We then train each AI-based 3D reconstruction model on the training set and evaluate their rendering quality on the test images. Table~\ref{fig:image_metrics} presents the results of this experiment.

\paragraph{\textbf{Evaluation Protocol:}} 
The evaluation protocol involves an initial coarse alignment of the 3D reconstructions with the ground-truth acquired data. Specifically, ArUco markers are placed in each scene to retrieve a global transformation and scale the 3D reconstruction for evaluation. We emphasize that these markers are not utilized for pose estimation or any other process by the evaluated methods.
A common next step is applying the Iterative Closest Point algorithm (ICP)~\cite{rusinkiewicz2001efficient} for fine alignment. 
However, the convergence of the ICP algorithm can be affected by errors in the reconstructed geometry. Some variations of the algorithm have been proposed to address these issues~\cite{rusinkiewicz2001efficient}, but these heuristics are often difficult to implement robustly across all evaluations.
%
Consequently, we opted to conduct manual alignment using 3D point cloud and mesh processing software. To ensure a fair and transparent evaluation, we provide the manually aligned meshes alongside the datasets.

\paragraph{\textbf{Metrics:}} 
To ensure that our evaluation concentrates solely on the wound area, we define a polygon around the wound in both the reconstructed mesh and the ground-truth point cloud. Within this polygon, we compute the following 3D reconstruction metrics: 
Average Distance (AD), Earth mover's Distance ($\mathcal{W}_2$), Hausdorff Distance (at percentiles $100^{th}$ (HD) and $90^{th}$ (HD$_{90}$)), and Normal Consistency (NC). 
We also leverage the optimal transport couplings computed for $\mathcal{W}_2$ to measure the normal consistency metric under optimal transport's assignment ($\mathcal{W}_2$-NC). 
The metrics AD, $\mathcal{W}_2$, and HD variants are measured in millimeters, while NC and $\mathcal{W}_2$-NC assess the alignment between reconstructed surface normals and ground-truth normals, achieving a maximum value of one for perfect alignment. 
Finally, for assessing rendering quality, we adopt standard metrics from the NeRF literature: Peak Signal-to-Noise Ratio (PSNR), Structural Similarity Index Measure (SSIM), and Learned Perceptual Image Patch Similarity (LPIPS).

Additionally, we report in Table~\ref{tab:acc_resources} reconstruction time (t), CPU and GPU memory usage of each method. Our experiments were conducted on an HPC node equipped with 16 Intel Xeon Platinum 8452Y CPUs, an NVIDIA H100 GPU, and 32GB of RAM.

\section{Results \& Discussion}

\subsection{Accuracy Evaluation}

As shown in Table~\ref{tab:acc_resources}, Meshroom, Neuralangelo, and Neus-facto consistently outperform other methods in the accuracy evaluation across various metrics, wound types, and recording devices. In particular, Neuralangelo and Neus-facto excel with complex wound geometries such as SD and PIS4. Additionally, Neus-facto exhibits robustness to variations in recording device quality. Even with Logitech videos more prone to blurriness and with lower resolution compared to iPhone videos, Neus-facto consistently produces good results. It particularly excels in NC and $\mathcal{W}_2$-NC, highlighting the benefit of the SDF formulation in generating smooth reconstructed surfaces.

In contrast, Gaussian Splatting methods struggle with sparsely observed regions, inheriting this limitation from the original 3DGS work. Given our dataset's realistic scenario design, these methods failed to achieve sufficient accuracy, a finding also reflected in the qualitative assessment of novel-view renderings.

Regarding fast reconstruction methods such as InstantNGP, NeuS2, and DUSt3R, the generated meshes exhibit poor quality, containing numerous floaters and defects that do not meet our application's requirements. Therefore, we have omitted their results from the main manuscript, but we provide a qualitative discussion of their results in the supplementary materials. Additionally, the rendering results of InstantNGP are reported in Table~\ref{fig:image_metrics}.

\begin{figure*}[t!]
    \centering
    \includegraphics[width=0.98\linewidth]{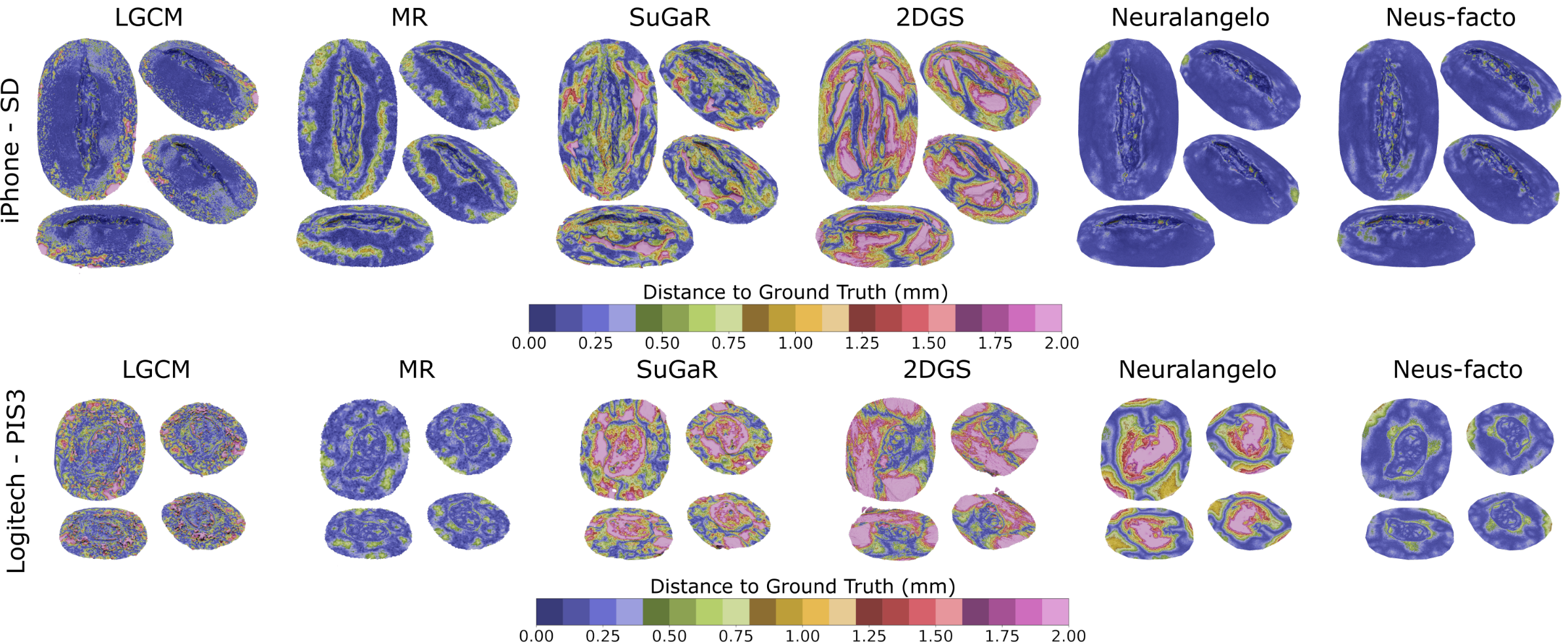}
    
    \caption{Figure depicting the reconstructed surfaces, where the color represents the distance to the closest point in the ground-truth point cloud, obtained from iPhone and Logitech recordings of SD and PIS3 respectively. Additional results for different wounds and devices are available in the supplementary materials.}
    \label{fig:error_coded_surfaces}
\vspace{-6pt}
\end{figure*}

\subsection{Precision Evaluation}


In Table~\ref{tab:prec_new}, we examine the repeatability of the reconstructions for the most accurate methods identified in the previous experiment (\textit{i.e.},~Meshroom, Neuralangelo, and Neus-facto) between different camera devices (inter-device column) and recording attempts (inter-recording column). 
Both Neuralangelo and Neus-facto demonstrate advantages over Meshroom (the most accurate photogrammetry method in our evaluation).
Remarkably, Neus-facto exhibits the best performance in the inter-device evaluation, showing the most consistent reconstructions among the compared methods.
%

\subsection{Qualitative Analysis}
We conduct a qualitative analysis of the evaluated models using error color-coded surfaces. In Figure~\ref{fig:error_coded_surfaces}, it appears that photogrammetric pipelines often yield noisy surfaces, primarily due to challenges in stereo matching under uncontrolled illumination in the videos. 
Similarly, Gaussian splatting approaches encounter similar difficulties due to the Gaussians' initialization based on the SfM point cloud. While these points reflect the challenges associated with stereo-matching, Gaussian splatting\textemdash which relies on explicit modelling of the neural surface representation\textemdash becomes significantly affected by noise introduced in the photogrammetry pipelines.
In contrast, Neuralanglo and Neus-facto leverage implicit volumetric rendering based on the SDF representation of a regular surface, allowing them to regularize noise that may exist in the training data. This results in accurate, repeatable, and smooth reconstructions.

\begin{figure}
    \centering
    \includegraphics[width=0.4\textwidth]{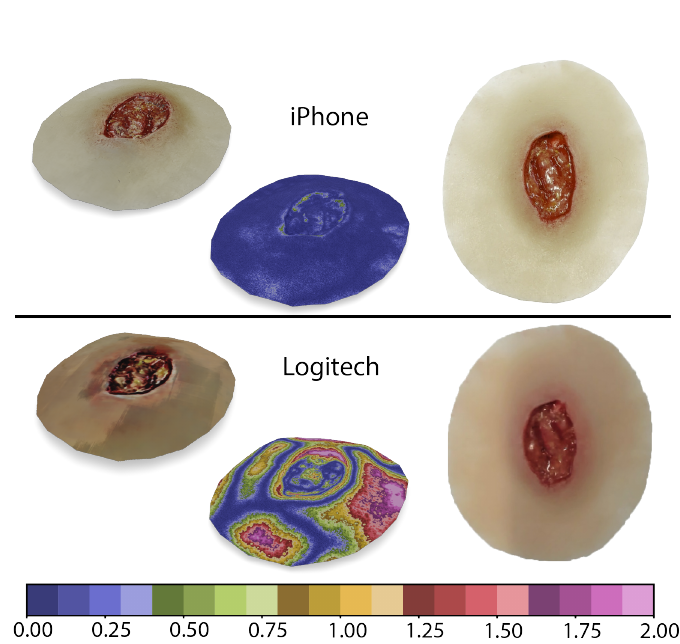}
    \caption{Figure illustrating the influence of image quality on Neuralangelo's performance. The best reconstruction was obtained with iPhone acquisitions, which feature 4K resolution compared to Logitech's lower image resolution and quality. 
    From left to right: reconstructed mesh, error color-coded mesh, a sample image of the wound sourced from the training set.}
    \label{fig:NABadLowRes}
\vspace{-6pt}
\end{figure}

Reproducible performance across different imaging devices is a desirable property for telehealth applications. As detailed in Table~\ref{tab:acc_resources} and depicted in Figure~\ref{fig:NABadLowRes}, NeuralAngelo exhibits a massive performance drop when the quality of the input images is reduced, we attribute this behavior to the model's large size and the reduced photogrammetric consistency between view acquired with a low-resolution device.

\input{tables/image_metrics}

\begin{figure*}
    \centering
    \includegraphics[width=0.98\linewidth]{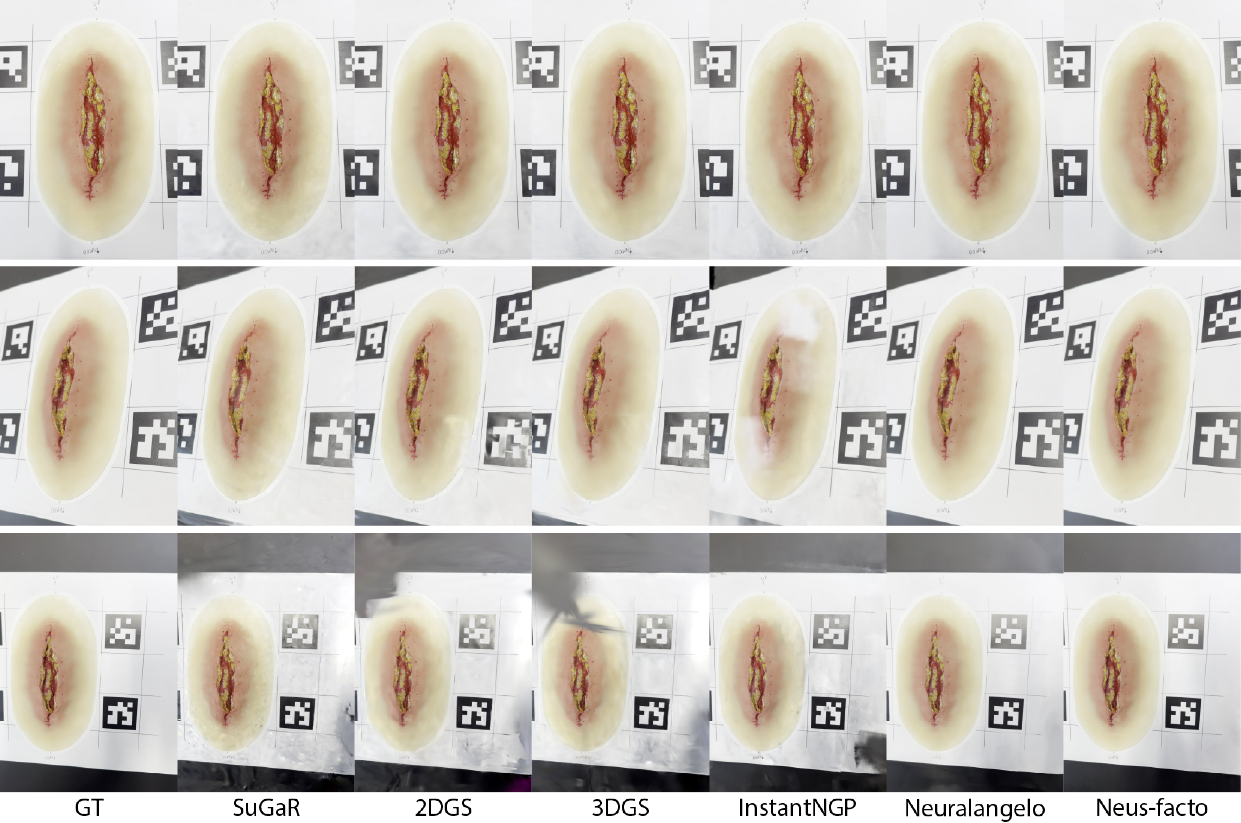}
    \caption{Qualitative evaluation of rendering methods for the SD wound model. 
    The first row depicts an image in the test set that is well-represented in the training set. The second row shows a rendering of the wound from an oblique view, where non-implicit methods may exhibit degraded performance (presence of floaters). 
    The last row displays an image far from the training view distribution, where the rendering can be challenging due to the scarcity of training views.}
    \label{fig:ComparisonRendering}
\end{figure*}

\subsection{Rendering Quality}
In Table~\ref{fig:image_metrics}, Neuralangelo and Neus-facto significantly outperform Instant-NGP (SOTA approach for novel view rendering) in terms of image rendering quality. This superior performance can be attributed to their ability to accurately represent surfaces, mitigating the floaters commonly found in Instant-NGP and 3DGS, and reducing rendering artifacts present in novel images that are too different from the training distribution (see Figure~\ref{fig:ComparisonRendering}). 
Despite Instant-NGP's fast runtime, it was excluded from the quantitative and qualitative evaluations due to its inability to produce surface meshes of good quality. Consequently, Neus-facto emerged as the overall best, balancing the trade-off between reconstruction quality, repeatability, and manageable training time.
Rendered images and additional qualitative results are made available in the supplementary materials. 



\section{Conclusion}

This paper presents a comprehensive study on 3D wound reconstruction using consumer-grade videos. We introduce \NAME, a dataset for 3D wound reconstruction and visualization, featuring video recordings of three wound types captured with two different cameras. These videos include photometric challenges typical in clinical settings, such as transient shadows, motion blur, and specular reflections. \NAME~ also provides accurate 3D ground truth acquired with a commercial 3D scanner and a rigorous evaluation protocol for assessing the geometric accuracy and precision of 3D reconstruction methods. Using this dataset, we explore a wide array of 3D reconstruction methods, from state-of-the-art neural rendering to traditional photogrammetry. We believe this study and dataset are crucial for advancing the use of this technology in wound care practices.

Beyond its clinical significance, this study demonstrates some limitations of current AI-based approaches, such as repeatability issues with SfM-initialized methods and the reduced quality of NeuralAngelo's 3D reconstruction under degraded image quality. We believe that these findings and benchmarks will also serve as valuable resources for the computer vision community.

\paragraph{\textbf{Acknowledgments:}} This work was supported by the
MRFF Rapid Applied Research Translation grant
(RARUR000158), CSIRO AI4M Minimising Antimicrobial Resistance Mission, and Australian Government
Training Research Program (AGRTP) Scholarship.

\paragraph{\textbf{Compliance with Ethical Standards:}} This study
was performed in line with the principles of the Declaration of Helsinki. The experimental procedures
involving human subjects described in this paper
were approved by CSIRO Health and Medical Human Research Ethics Committee (CHMHREC) [ethics protocol 2022\_015\_LR].


{\small
\bibliographystyle{ieee_fullname}
\bibliography{egbib}
}

\include{supplementary/supplementary}

\end{document}

%% file: tables/accuracy_plus_resources.tex
\begin{table*}[!htb]\centering
\scriptsize
\begin{adjustbox}{width=2\columnwidth,center}
\begin{tabular}{crccccccccccccccccccc}\toprule

\multicolumn{1}{c}{} &\multicolumn{1}{c}{} &\multicolumn{18}{c}{\textbf{Accuracy}} \\
\cmidrule(lr){3-20}

\multicolumn{1}{c}{} &\multicolumn{1}{c}{} &\multicolumn{9}{c}{\textbf{iPhone}} &\multicolumn{9}{c}{\textbf{Logitech}} \\
\cmidrule(lr){3-11}
\cmidrule(lr){12-20}

\multicolumn{1}{c}{\textbf{Wound}} &\multicolumn{1}{c}{\textbf{Method}} &\textbf{AD} &$\mathcal{W}_2$ &\textbf{HD} &\textbf{HD}$_{90}$ &\textbf{NC} &$\mathcal{W}_2$-\textbf{NC} &\textbf{t} &\textbf{CPU} &\textbf{GPU} &\textbf{AD} &$\mathcal{W}_2$ &\textbf{HD} &\textbf{HD}$_{90}$ &\textbf{NC} &$\mathcal{W}_2$-\textbf{NC} &\textbf{t} &\textbf{CPU} &\textbf{GPU} \\\midrule

\multirow{7}{*}{PIS3} &CM &0.305 &1.702 &2.757 &0.624 &0.812 &0.797 &1.26h &10.1GB &\cellcolor{2}4.73GB &\cellcolor{3}0.696 &3.106 &14.775 &2.131 &0.745 &0.564 &\cellcolor{3}31.2m &\cellcolor{2}2.8GB &\cellcolor{2}2GB \\
&LGCM &\cellcolor{3}0.259 &\cellcolor{3}1.632 &2.858 &\cellcolor{3}0.516 &0.830 &0.823 &1.29h &11GB &14.1GB &0.751 &4.470 &12.131 &2.653 &0.739 &0.549 &31.7m &\cellcolor{1}\textbf{2.7GB} &\cellcolor{3}4.4GB \\
&*MR &\cellcolor{1}\textbf{0.230} &\cellcolor{1}\textbf{1.575} &\cellcolor{1}\textbf{1.272} &\cellcolor{1}\textbf{0.454} &\cellcolor{2}0.992 &\cellcolor{3}0.991 &\cellcolor{1}\textbf{15.3m} &\cellcolor{3}8GB &\cellcolor{1}\textbf{1.3GB} &\cellcolor{1}\textbf{0.249} &\cellcolor{1}\textbf{1.572} &79.029 &\cellcolor{1}\textbf{0.504} &\cellcolor{2}0.990 &\cellcolor{2}0.990 &\cellcolor{1}\textbf{4.3m} &4.7GB &\cellcolor{1}\textbf{0.7GB} \\
&SuGaR &0.806 &4.011 &9.657 &1.746 &0.906 &0.822 &\cellcolor{3}1.23h &\cellcolor{1}\textbf{5.7GB} &71.1GB &1.075 &2.336 &\cellcolor{2}6.317 &2.342 &0.898 &0.869 &45.6m &6GB &20.4GB \\
&2DGS &0.757 &3.069 &6.730 &1.695 &0.726 &0.682 &1.66h &23.5GB &31.4GB &1.402 &3.826 &\cellcolor{3}7.380 &3.170 &0.789 &0.759 &\cellcolor{2}27.6m &19.6GB &7.8GB \\
&Neuralangelo &\cellcolor{2}0.257 &\cellcolor{2}1.614 &\cellcolor{2}1.359 &\cellcolor{2}0.506 &\cellcolor{2}0.992 &\cellcolor{2}0.992 &15.54h &\cellcolor{2}7.1GB &\cellcolor{3}9.7GB &0.982 &\cellcolor{3}1.990 &14.605 &\cellcolor{3}2.123 &\cellcolor{3}0.977 &\cellcolor{3}0.967 &8.17h &\cellcolor{3}4GB &9.7GB \\
&Neus-facto &0.282 &1.669 &\cellcolor{3}1.681 &0.569 &\cellcolor{1}\textbf{0.993} &\cellcolor{1}\textbf{0.993} &\cellcolor{2}55.4m &19GB &24.3GB &\cellcolor{2}0.263 &\cellcolor{2}1.612 &\cellcolor{1}\textbf{1.632} &\cellcolor{2}0.621 &\cellcolor{1}\textbf{0.993} &\cellcolor{1}\textbf{0.993} &46.8m &12.2GB &13.9GB \\
\cmidrule(lr){1-20}

\multirow{7}{*}{PIS4} &CM &\cellcolor{3}0.215 &1.572 &2.689 &\cellcolor{3}0.499 &0.853 &0.843 &1.27h &10.3GB &\cellcolor{2}4.7GB &\cellcolor{3}0.827 &3.668 &12.438 &2.262 &0.806 &0.697 &\cellcolor{3}31m &\cellcolor{1}\textbf{2.8GB} &\cellcolor{2}2GB \\
&LGCM &0.216 &1.766 &\cellcolor{2}2.313 &0.503 &0.868 &0.859 &1.3h &10.5GB &14.2GB &1.068 &6.638 &19.263 &3.874 &0.766 &0.559 &32.7m &\cellcolor{2}3.5GB &\cellcolor{3}4.4GB \\
&*MR &\cellcolor{2}0.192 &\cellcolor{3}1.556 &2.664 &\cellcolor{2}0.475 &\cellcolor{2}0.991 &\cellcolor{3}0.984 &\cellcolor{1}\textbf{15.8m} &\cellcolor{2}7.7GB &\cellcolor{1}\textbf{1.3GB} &\cellcolor{2}0.424 &\cellcolor{2}1.779 &\cellcolor{3}3.503 &\cellcolor{2}1.030 &\cellcolor{2}0.982 &\cellcolor{2}0.980 &\cellcolor{1}\textbf{4.4m} &4.7GB &\cellcolor{1}\textbf{0.7GB} \\
&SuGaR &0.657 &2.003 &4.608 &1.437 &0.950 &0.942 &\cellcolor{3}1.2h &\cellcolor{1}\textbf{5.1GB} &61.6GB &3.342 &7.144 &17.882 &10.957 &0.709 &0.462 &42.38m &6GB &21.5GB \\
&2DGS &0.747 &2.570 &6.364 &1.835 &0.833 &0.808 &1.61h &23.5GB &24.2GB &1.057 &2.623 &7.203 &2.409 &0.858 &0.846 &\cellcolor{2}27.3m &16.2GB &5GB \\
&Neuralangelo &\cellcolor{1}\textbf{0.178} &\cellcolor{1}\textbf{1.546} &\cellcolor{3}2.472 &\cellcolor{1}\textbf{0.427} &\cellcolor{1}\textbf{0.992} &\cellcolor{1}\textbf{0.985} &16.37h &\cellcolor{2}7.7GB &\cellcolor{3}9.7GB &0.953 &\cellcolor{3}2.076 &\cellcolor{2}3.165 &\cellcolor{3}2.006 &\cellcolor{3}0.972 &\cellcolor{3}0.975 &8.15h &\cellcolor{3}4.3GB &9.7GB \\
&Neus-facto &0.217 &\cellcolor{2}1.555 &\cellcolor{1}\textbf{2.301} &0.539 &\cellcolor{2}0.991 &\cellcolor{1}\textbf{0.985} &\cellcolor{2}51.8m &19GB &24.2GB &\cellcolor{1}\textbf{0.256} &\cellcolor{1}\textbf{1.607} &\cellcolor{1}\textbf{2.567} &\cellcolor{1}\textbf{0.639} &\cellcolor{1}\textbf{0.991} &\cellcolor{1}\textbf{0.985} &47.6m &12.2GB &14GB \\
\cmidrule(lr){1-20}

\multirow{7}{*}{SD} &CM &\cellcolor{3}0.355 &7.547 &9.956 &0.730 &0.887 &0.817 &\cellcolor{3}1.2h &10GB &\cellcolor{2}4.7GB &0.963 &8.467 &26.819 &2.962 &0.775 &0.578 &\cellcolor{3}32.7m &\cellcolor{1}\textbf{3GB} &\cellcolor{2}2GB \\
&LGCM &0.366 &\cellcolor{3}7.297 &16.488 &\cellcolor{3}0.722 &0.892 &0.819 &1.3h &10.5GB &14.1GB &\cellcolor{3}0.595 &\cellcolor{3}7.639 &23.129 &\cellcolor{3}1.648 &0.828 &0.687 &33.7m &\cellcolor{1}\textbf{3GB} &\cellcolor{3}4.4GB \\
&*MR &0.357 &7.765 &\cellcolor{3}3.292 &0.836 &\cellcolor{3}0.979 &\cellcolor{2}0.924 &\cellcolor{1}\textbf{13.9m} &\cellcolor{3}7.2GB &\cellcolor{1}\textbf{1.3GB} &0.736 &7.701 &\cellcolor{3}5.767 &1.866 &\cellcolor{3}0.956 &\cellcolor{3}0.933 &\cellcolor{1}\textbf{4.5m} &\cellcolor{3}4.7GB &\cellcolor{1}\textbf{0.7GB} \\
&SuGaR &0.698 &8.455 &9.625 &1.559 &0.932 &0.823 &1.25h &\cellcolor{1}\textbf{5.6GB} &94GB &2.295 &11.394 &15.475 &6.473 &0.767 &0.597 &43.9m &5.9GB &19.5GB \\
&2DGS &1.195 &8.315 &8.274 &2.484 &0.872 &0.830 &1.65h &23.5GB &24.6GB &1.846 &9.076 &10.209 &3.818 &0.849 &0.804 &\cellcolor{2}26.2m &13GB &5.5GB \\
&Neuralangelo &\cellcolor{1}\textbf{0.201} &\cellcolor{1}\textbf{7.149} &\cellcolor{2}3.115 &\cellcolor{1}\textbf{0.439} &\cellcolor{2}0.980 &\cellcolor{3}0.906 &14.23h &\cellcolor{1}\textbf{5.6GB} &\cellcolor{3}9.7GB &\cellcolor{2}0.547 &\cellcolor{2}7.460 &\cellcolor{2}5.529 &\cellcolor{2}1.366 &\cellcolor{2}0.957 &\cellcolor{2}0.935 &8.1h &5GB &9.7GB \\
&Neus-facto &\cellcolor{2}0.230 &\cellcolor{2}7.257 &\cellcolor{1}\textbf{3.048} &\cellcolor{2}0.510 &\cellcolor{1}\textbf{0.986} &\cellcolor{1}\textbf{0.925} &\cellcolor{2}52.5m &19GB &24.2GB &\cellcolor{1}\textbf{0.428} &\cellcolor{1}\textbf{7.192} &\cellcolor{1}\textbf{5.045} &\cellcolor{1}\textbf{1.034} &\cellcolor{1}\textbf{0.975} &\cellcolor{1}\textbf{0.939} &46.2m &12.2GB &14GB \\
\bottomrule
\end{tabular}
\end{adjustbox}
\caption{Table presenting the accuracy of different 3D reconstruction methods across three wound types (PIS3, PIS4, and SD) and two acquisition devices (iPhone and Logitech). Results highlight the superiority of Meshroom (MR), Neuralangelo, and Neus-facto across six metrics evaluated: AD, $\mathcal{W}_2$, HD, and $\text{HD}_{90}$ (reported in millimetres), along with NC and $\mathcal{W}_2$-NC values (ranging from 0 for orthogonal to 1 for the same orientation). Meshroom emerges for best time complexity (t), while SuGaR and 2DGS require the most memory.*Due to compatibility issues during the setup of Meshroom with our H100 GPUs we used an HPC node with an NVIDIA RTX A6000 instead.}
\label{tab:acc_resources}
\end{table*}

%% file: tables/precision_new.tex
\begin{table*}[!htb]\centering
\caption{Table presenting the precision of the best methods from the accuracy experiment. We report precision for inter-recording in the iPhone acquisitions and inter-device across the three wound types. Results highlight Neus-facto as the most robust method among those evaluated.}\label{tab:prec_new}
\begin{adjustbox}{width=1.75\columnwidth,center}
\scriptsize
\begin{tabular}{crccccccccccccc}\toprule

\multicolumn{1}{c}{} &\multicolumn{1}{c}{} &\multicolumn{12}{c}{\textbf{Precision}} \\
\cmidrule(lr){3-14}

\multicolumn{1}{c}{\textbf{}} &\multicolumn{1}{c}{} &\multicolumn{6}{c}{\textbf{Inter recording (iPhone)}} &\multicolumn{6}{c}{\textbf{Inter device}} \\
\cmidrule(lr){3-8}
\cmidrule(lr){9-14}
\multicolumn{1}{c}{\textbf{Wound}} &\multicolumn{1}{c}{\textbf{Method}} &\textbf{AD} &$\mathcal{W}_2$ &\textbf{HD} &\textbf{HD}$_{90}$ &\textbf{NC} &$\mathcal{W}_2$-\textbf{NC} &\textbf{AD} &$\mathcal{W}_2$ &\textbf{HD} &\textbf{HD}$_{90}$ &\textbf{NC} &$\mathcal{W}_2$-\textbf{NC} \\\midrule

\multirow{3}{*}{PIS3} &MR &\cellcolor{1}\textbf{0.089} &0.671 &2.099 &\cellcolor{1}\textbf{0.138} &0.994 &0.970 &\cellcolor{3}0.224 &\cellcolor{3}0.617 &\cellcolor{3}2.266 &\cellcolor{3}0.441 &\cellcolor{3}0.991 &\cellcolor{3}0.979 \\
&Neuralangelo &\cellcolor{3}0.102 &\cellcolor{1}\textbf{0.405} &\cellcolor{1}\textbf{0.452} &\cellcolor{3}0.183 &\cellcolor{1}\textbf{0.997} &\cellcolor{1}\textbf{0.997} &0.865 &1.578 &9.191 &1.611 &0.981 &0.966 \\
&Neus-facto &0.126 &\cellcolor{3}0.457 &\cellcolor{3}0.476 &0.229 &\cellcolor{1}\textbf{0.997} &\cellcolor{1}\textbf{0.997} &\cellcolor{1}\textbf{0.197} &\cellcolor{1}\textbf{0.493} &\cellcolor{1}\textbf{0.854} &\cellcolor{1}\textbf{0.382} &\cellcolor{1}\textbf{0.998} &\cellcolor{1}\textbf{0.998} \\
\cmidrule(lr){1-14}

\multirow{3}{*}{PIS4} &MR &0.216 &0.542 &1.990 &0.499 &0.988 &0.989 &\cellcolor{3}0.441 &\cellcolor{3}0.736 &\cellcolor{3}2.646 &\cellcolor{3}0.946 &\cellcolor{3}0.984 &\cellcolor{3}0.984 \\
&Neuralangelo &\cellcolor{1}\textbf{0.093} &\cellcolor{1}\textbf{0.443} &\cellcolor{3}1.382 &\cellcolor{1}\textbf{0.160} &\cellcolor{3}0.994 &\cellcolor{3}0.992 &0.645 &1.000 &3.256 &1.482 &0.983 &0.982 \\
&Neus-facto &\cellcolor{3}0.099 &\cellcolor{3}0.449 &\cellcolor{1}\textbf{0.593} &\cellcolor{3}0.176 &\cellcolor{1}\textbf{0.995} &\cellcolor{1}\textbf{0.994} &\cellcolor{1}\textbf{0.286} &\cellcolor{1}\textbf{0.561} &\cellcolor{1}\textbf{1.556} &\cellcolor{1}\textbf{0.548} &\cellcolor{1}\textbf{0.992} &\cellcolor{1}\textbf{0.992} \\
\cmidrule(lr){1-14}

\multirow{3}{*}{SD} &MR &0.347 &\cellcolor{1}\textbf{0.826} &5.954 &0.710 &0.978 &\cellcolor{3}0.976 &0.663 &1.676 &5.041 &1.449 &0.961 &0.952 \\
&Neuralangelo &\cellcolor{1}\textbf{0.279} &\cellcolor{3}0.888 &\cellcolor{1}\textbf{2.078} &\cellcolor{1}\textbf{0.533} &\cellcolor{3}0.984 &0.974 &\cellcolor{1}\textbf{0.330} &\cellcolor{3}0.916 &\cellcolor{1}\textbf{2.363} &\cellcolor{1}\textbf{0.657} &\cellcolor{3}0.979 &\cellcolor{3}0.973 \\
&Neus-facto &\cellcolor{1}\textbf{0.279} &0.935 &\cellcolor{3}2.122 &\cellcolor{3}0.608 &\cellcolor{1}\textbf{0.986} &\cellcolor{1}\textbf{0.987} &\cellcolor{3}0.346 &\cellcolor{1}\textbf{0.886} &\cellcolor{3}2.516 &\cellcolor{3}0.709 &\cellcolor{1}0.984 &\cellcolor{1}\textbf{0.984} \\

\bottomrule
\end{tabular}
\end{adjustbox}
\end{table*}

%% file: tables/image_metrics.tex
\begin{table}[!t]
    \centering
    \caption{Table presenting the averaged image metrics for the novel-view renderings experiment. We compare NeRF and Gaussian splatting methods for the iPhone acquisitions of PIS3, PIS4, and SD.}
    \begin{adjustbox}{width=\columnwidth,center}
    \begin{tabular}{rcccccccccc}\midrule
        \multicolumn{1}{c}{\textbf{Method}} &\multicolumn{3}{c}{\textbf{PSNR} $\uparrow$} &\multicolumn{3}{c}{\textbf{SSIM} $\uparrow$} &\multicolumn{3}{c}{\textbf{LPIPS} $\downarrow$} &\multicolumn{1}{c}{\textbf{Time 4k}} \\
        \cmidrule(lr){2-4}
        \cmidrule(lr){5-7}
        \cmidrule(lr){8-10}
        
        \multicolumn{1}{c}{} &\textbf{PIS3} &\textbf{PIS4} &\textbf{SD} &\textbf{PIS3} &\textbf{PIS4} &\textbf{SD} &\textbf{PIS3} &\textbf{PIS4} &\textbf{SD} &\textbf{(Mean)} \\
        \midrule
    
        SuGaR & 29.719 & 30.935 & 30.869 & 0.959 & 0.960 & 0.954 & 0.056 & 0.050 & 0.053 & 1.234h \\
        2DGS  & \cellcolor{3}30.874 & 31.489 & 31.852 & \cellcolor{3}0.966 & 0.965 & 0.959 & \cellcolor{3}0.055 & 0.050 & 0.050 & 1.695h \\
        3DGS  & 30.642 & \cellcolor{3}32.595 & \cellcolor{3}31.904 & \cellcolor{2}0.967 & \cellcolor{3}0.967 & \cellcolor{3}0.960 & \cellcolor{2}0.049 & \cellcolor{3}0.045 & \cellcolor{3}0.046 & 57.8m \\
        Instant-NGP  & 30.386 & 30.979 & 31.068 & 0.962 & 0.962 & 0.957 & 0.065 & 0.059 & 0.061 & 6.8m \\
        Neuralangelo  & \cellcolor{1}\textbf{33.897} & \cellcolor{1}\textbf{35.560} & \cellcolor{2}35.533 & \cellcolor{1}\textbf{0.971} & \cellcolor{1}\textbf{0.973} & \cellcolor{1}\textbf{0.967} & \cellcolor{1}\textbf{0.047} & \cellcolor{2}0.039 & \cellcolor{1}\textbf{0.038} & 17.884h \\
        Neus-facto & \cellcolor{2}32.224 & \cellcolor{2}35.096 & \cellcolor{1}\textbf{35.552} & 0.964 & \cellcolor{2}0.972 & \cellcolor{2}0.966 & 0.066 & \cellcolor{1}\textbf{0.038} & \cellcolor{1}\textbf{0.038} & 1.225h \\
        
        \bottomrule
    \end{tabular}
    \end{adjustbox}
    \label{fig:image_metrics}
\end{table}

%% file: supplementary/supplementary.tex
\clearpage
\setcounter{page}{1}
\renewcommand*{\thesection}{\Alph{section}}
\renewcommand*{\thesubsection}{\alph{subsection}}
\setcounter{section}{0}
\renewcommand\thefigure{\thesection.\arabic{figure}} 
\setcounter{figure}{0}

\twocolumn[{%
\renewcommand\twocolumn[1][]{#1}%
\maketitlesupplementary
\vspace{8pt}
}]

In this supplementary material, we present further details on the acquisition setup and additional qualitative results, including error color-coded meshes and novel-view renderings. Additionally, we present the non-suitability of the surfaces from DUSt3R~\cite{wang2023dust3rgeometric3dvision}, Instant-NGP~\cite{muller2022instant}, and NeuS2~\cite{wang2023neus2} for geometric assessment.

In Section~\ref{sec:a} we present more details relative to the \NAME's dataset challenges. 
Next, in Section~\ref{sec:b} we describe more precisely the evaluation protocol, namely a qualitative evaluation of the ground-truth point clouds and specifications about sampling of the meshes adopted for the experiments of Section 3.3 of the main paper.
Following,
with respect to Figure 4 and Figure 6 of the main paper,
we present the additional error color-coded meshes (Figure~\ref{fig:error_coded_surfaces}) and renderings (Figure~\ref{fig:renders_PIS3} for PIS3 and Figure~\ref{fig:renders_PIS4} for PIS4) for the remaining devices and wound types we omitted in the main paper.

In addition, we present the surfaces extracted from DUSt3R, Instant-NGP, and NeuS2 in Figures~\ref{fig:duster},~\ref{fig:ingp}, and~\ref{fig:nesu2} respectively, and justify why those methods are not included in 3D reconstruction benchmark.

Finally, in Figure~\ref{fig:higher_frames} we present some cases of failed reconstruction for the photogrammetric approaches when we input a greater amount of images, specifically 100 and 150. Compared to a set of 50 images, 100 and 150 present more blurring artefacts as mentioned in Section 3.1 of the main paper. As a result, we chose sets of 50 images for our \NAME~dataset as they provided more consistent results and better wound representations when compared to fewer images.

\section{Dataset additional challenges}\label{sec:a}

\begin{figure}[t!]
    \centering
    \includegraphics[width=\linewidth]{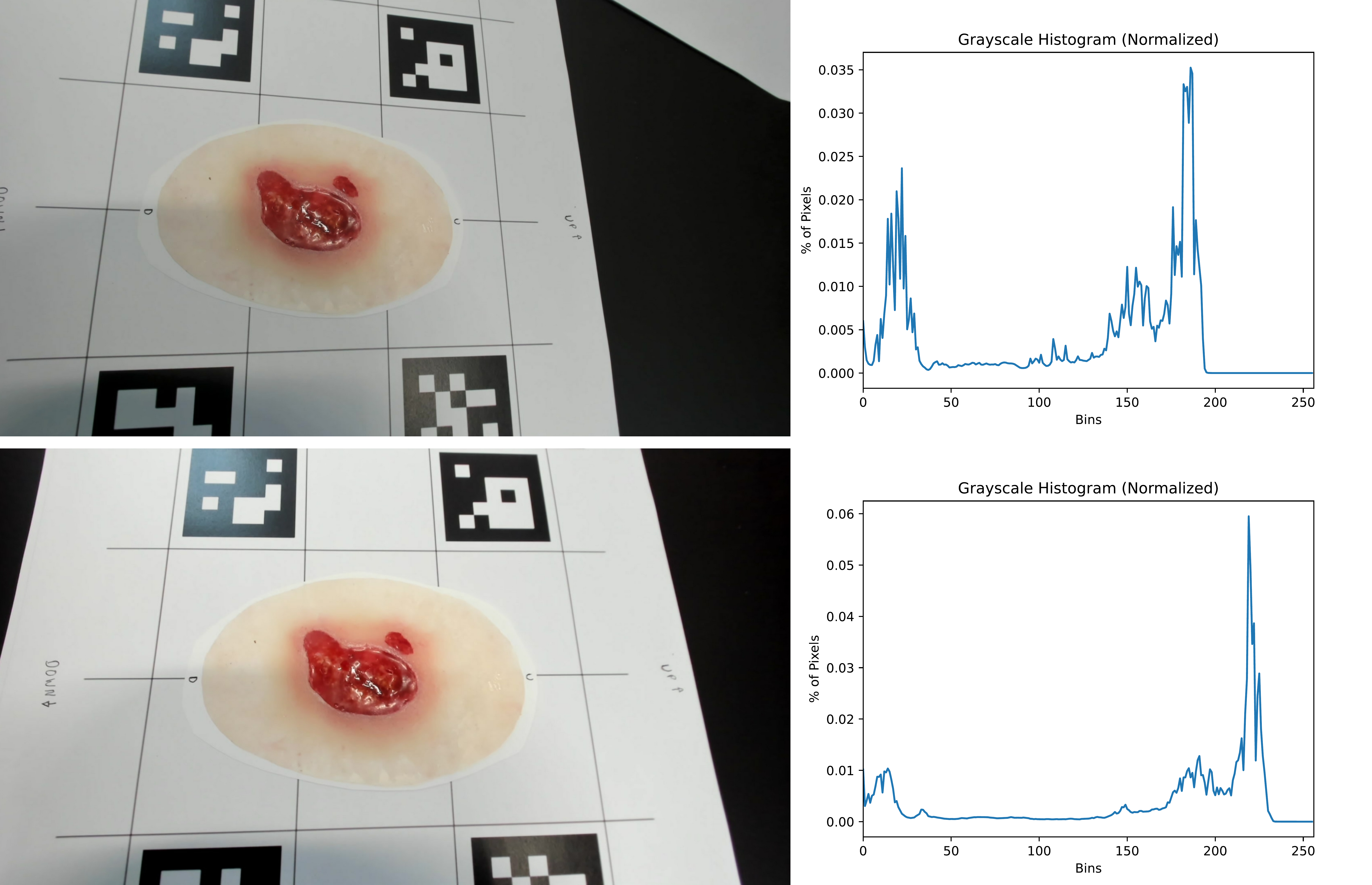}
    \caption{Two images of the PIS3 wound type captured from slightly different angles using the Logitech camera show different luminosity levels, as indicated by their normalized grayscale histograms (on the right). 
    }
    \label{fig:exp}
\end{figure}

As depicted in Figure~\ref{fig:setup} of the main paper, we use a Logitech 4K webcam and an iPhone 14 Pro Max for video recordings, and a Revopoint POP 3D scanner to acquire the ground-truth point clouds. Both recording devices utilise default acquisition settings (e.g. exposure time, frame rate, aperture, white balance, etc.) to simulate acquisition scenarios as close as possible to a telehealth application, Figure~\ref{fig:exp} shows an example of dynamic lighting conditions comprised in our dataset.
Furthermore, 
when extracting frames from the recordings, we select the sharpest frames as outlined in Section~\ref{sec:dataset} of the main paper, however, we might still observe blurred images in the datasets. An example regarding the Logitech recordings is partially observable from the image in the second row of Figure~\ref{fig:exp}.
Additionally, \NAME~presents moving shadows cast by the operator during the acquisition, which is a common issue in environments with multiple illumination sources.

\section{Evaluation details}\label{sec:b}

\begin{figure*}[t!]
    \centering
    \includegraphics[width=\linewidth]{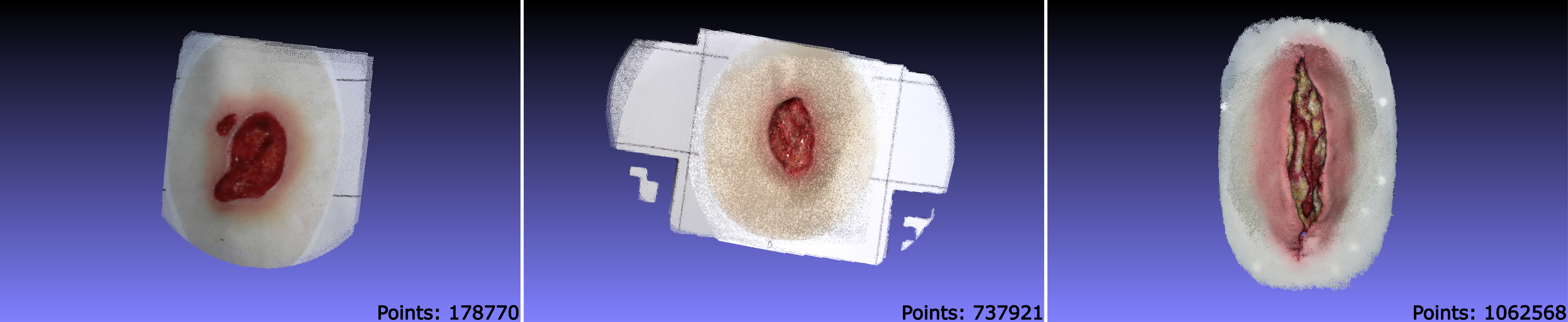}
    \caption{Ground-truth point clouds acquired with the Revopoint POP 3D scanner and their respective total number of points. From left to right: PIS3, PIS4, and SD wound types.}
    \label{fig:pclouds}
\end{figure*}

In Figure~\ref{fig:pclouds}, we illustrate qualitatively the three ground-truth point clouds acquired with the Revopoint POP 3D scanner.

For the metrics AD, HD, $\text{HD}_{90}$, and NC we uniformly sampled 2 million points from the reconstructed meshes, while for $\mathcal{W}_2$ and $\mathcal{W}_2$-NC we sampled around 30 thousand points from the reconstructed meshes due to the computational complexity of optimal transport metrics.

%
\begin{figure*}[t!]
    \centering
    \includegraphics[width=\linewidth]{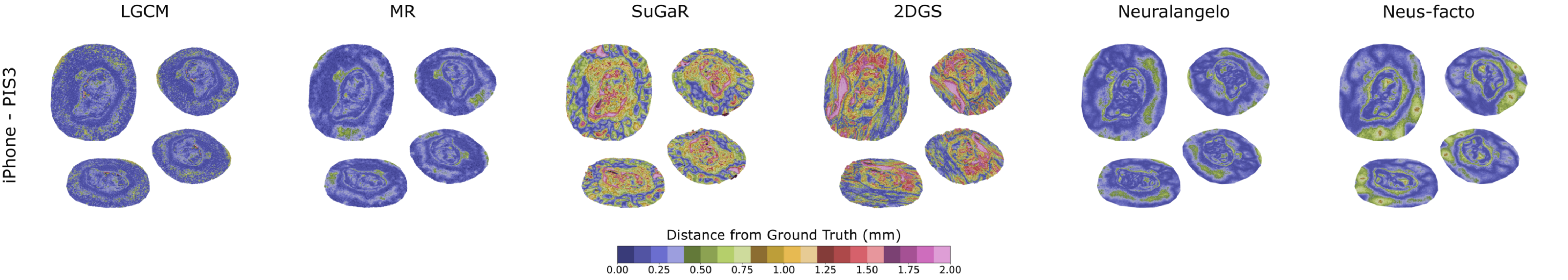}
    \includegraphics[width=\linewidth]{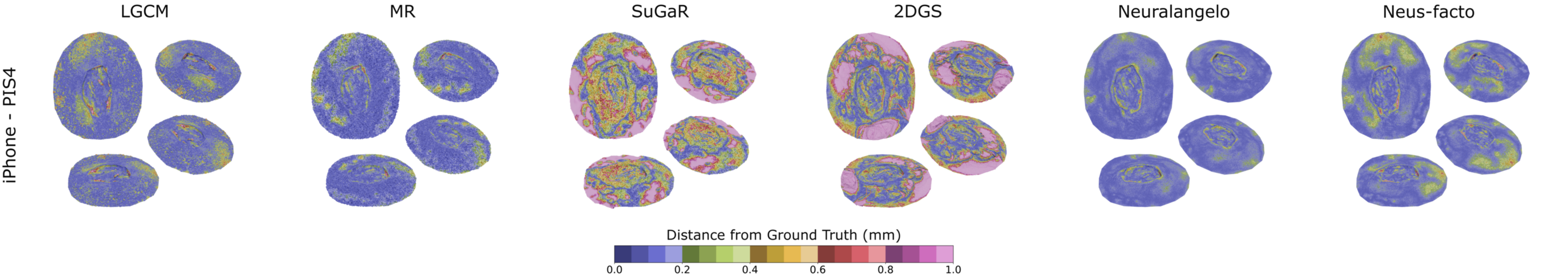}
    \includegraphics[width=\linewidth]{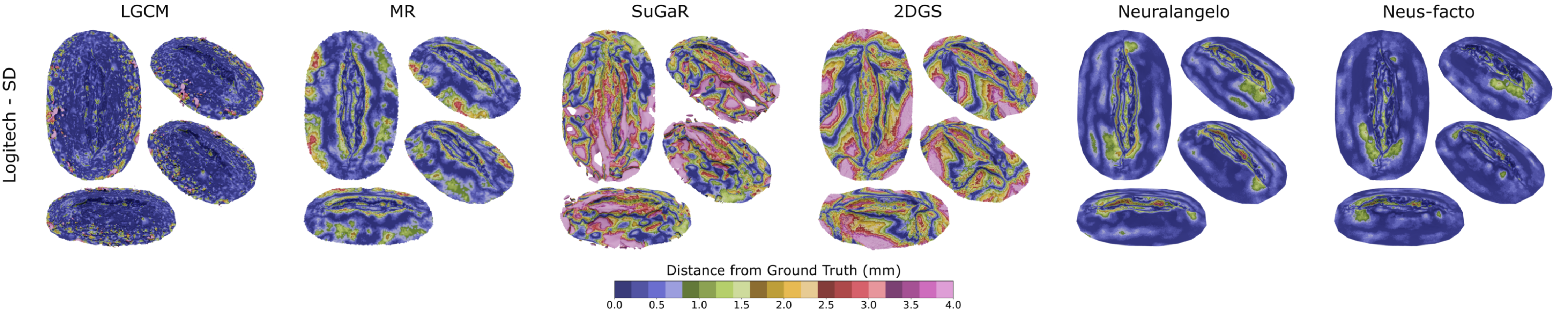}
    \includegraphics[width=\linewidth]{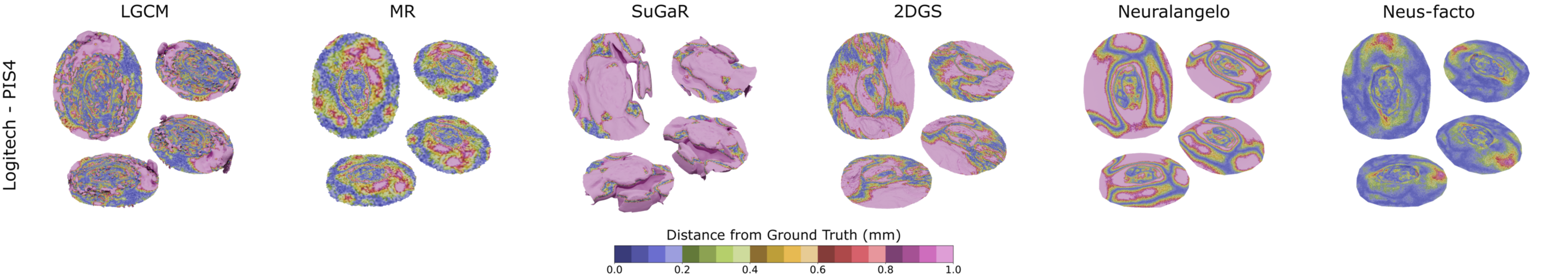}
    
    \caption{Reconstructed surface results for different methods, wound types and recording devices. The color represents the distance to the closest point in the ground-truth point cloud.}
    \label{fig:sup_error_coded_surfaces}
\vspace{-6pt}
\end{figure*}

\begin{figure*}[t!]
    \centering
    \includegraphics[width=\linewidth]{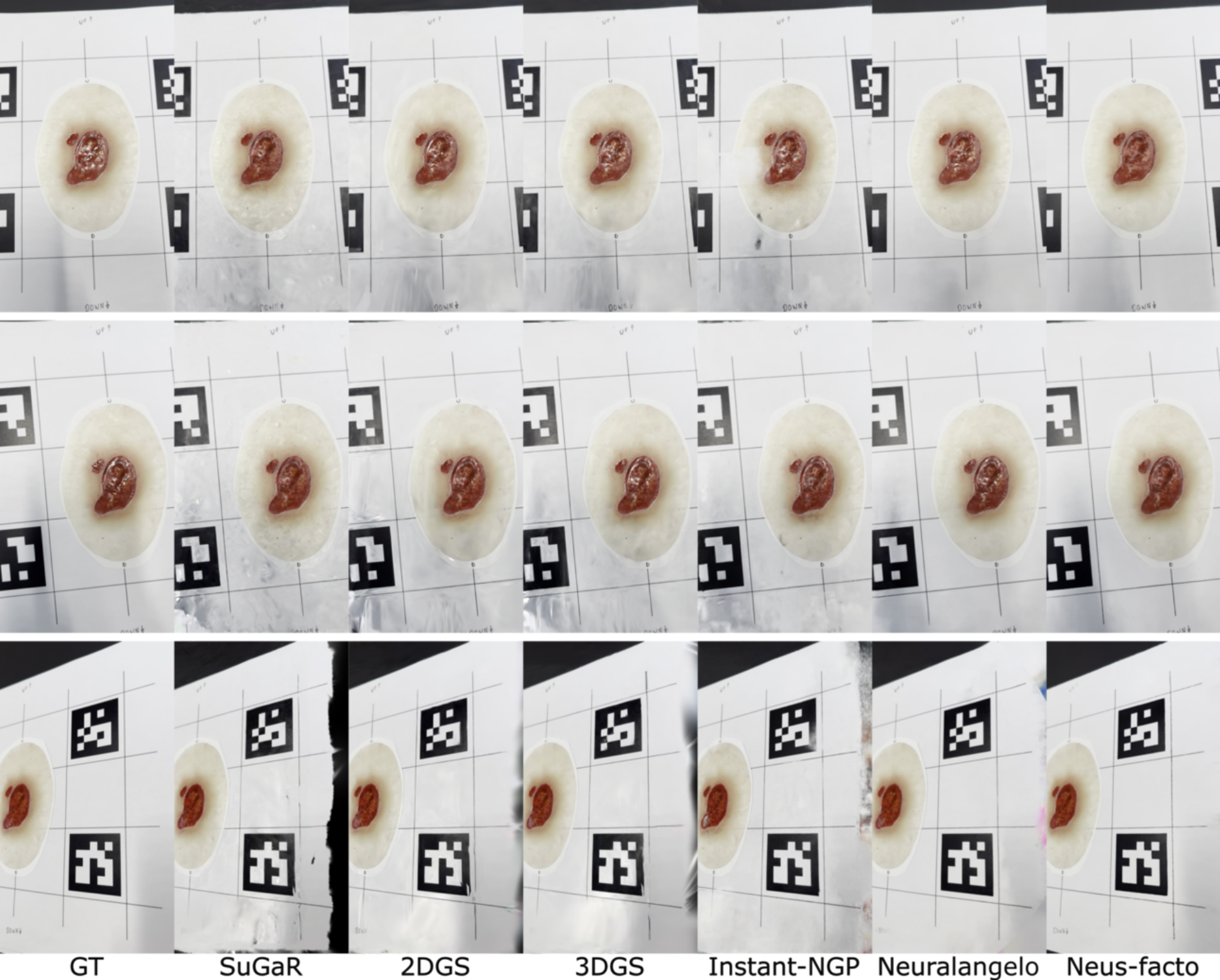}
    \caption{Qualitative evaluation of rendering methods for the PIS3 wound model. 
    The first row presents the renderings for a view perspective in the test set that is well-represented in the training set. The second row shows a rendering of the wound from an oblique view, where we can observe the presence of floaters in non-SDF-based methods. 
    The last row displays a view perspective under-observed in the training set. Notably, the PIS3 wound type presents hard reflections that appear especially in the first two rows.}
    \label{fig:renders_PIS3}
\end{figure*}

\begin{figure*}[t!]
    \centering
    \includegraphics[width=\linewidth]{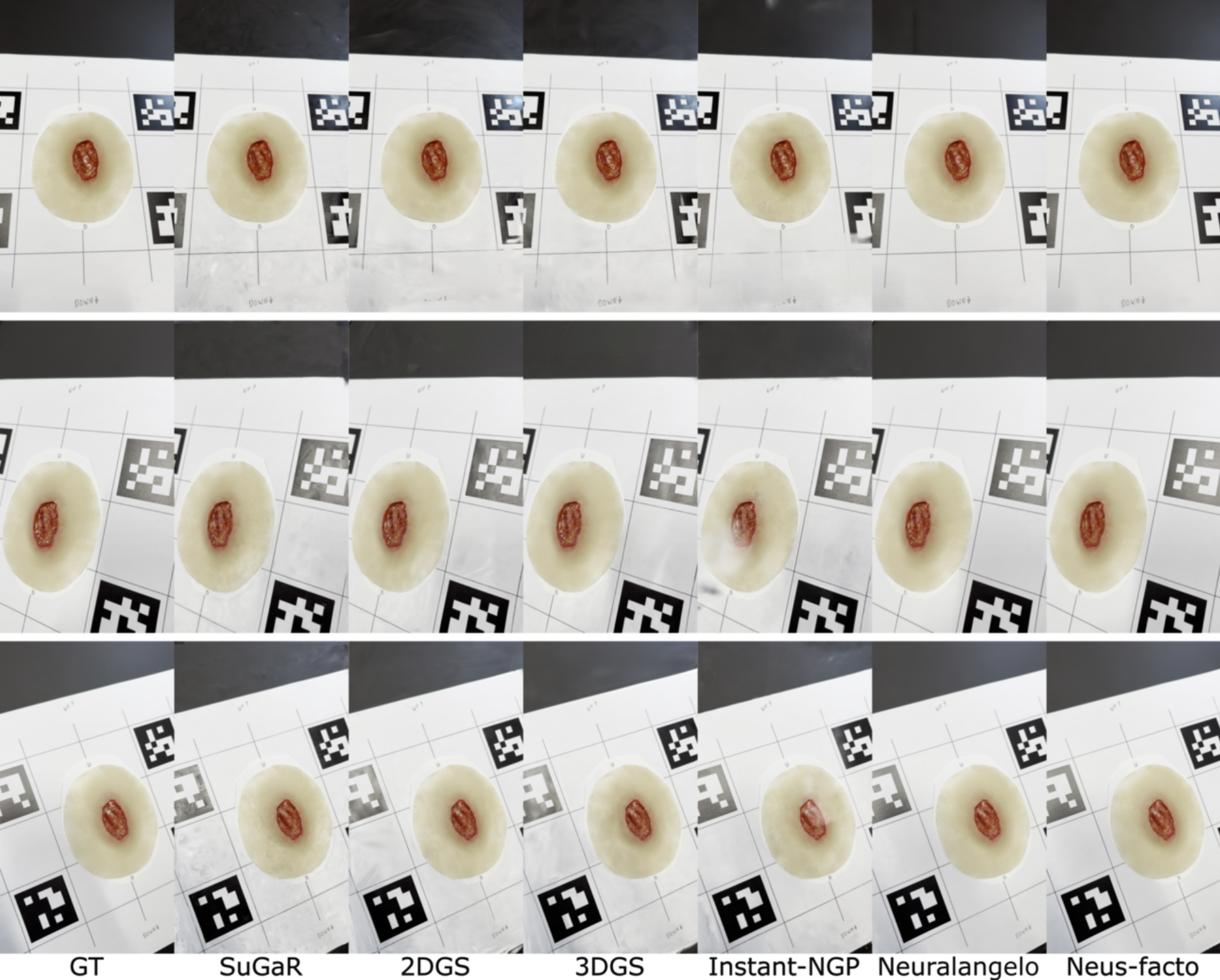}
    \caption{Qualitative evaluation of rendering methods for the PIS4 wound model. 
    The first row presents the renderings for a view perspective in the test set that is well-represented in training set, however, floaters are already visible for non-SDF-based methods. The second row shows a rendering of the wound from an oblique view, where more floaters are present. 
    The last row displays an image from another view angle where artifacts similar to those in the second row can be observed.}
    \label{fig:renders_PIS4}
\end{figure*}


\begin{figure*}[t!]
    \centering
    \includegraphics[width=\linewidth]{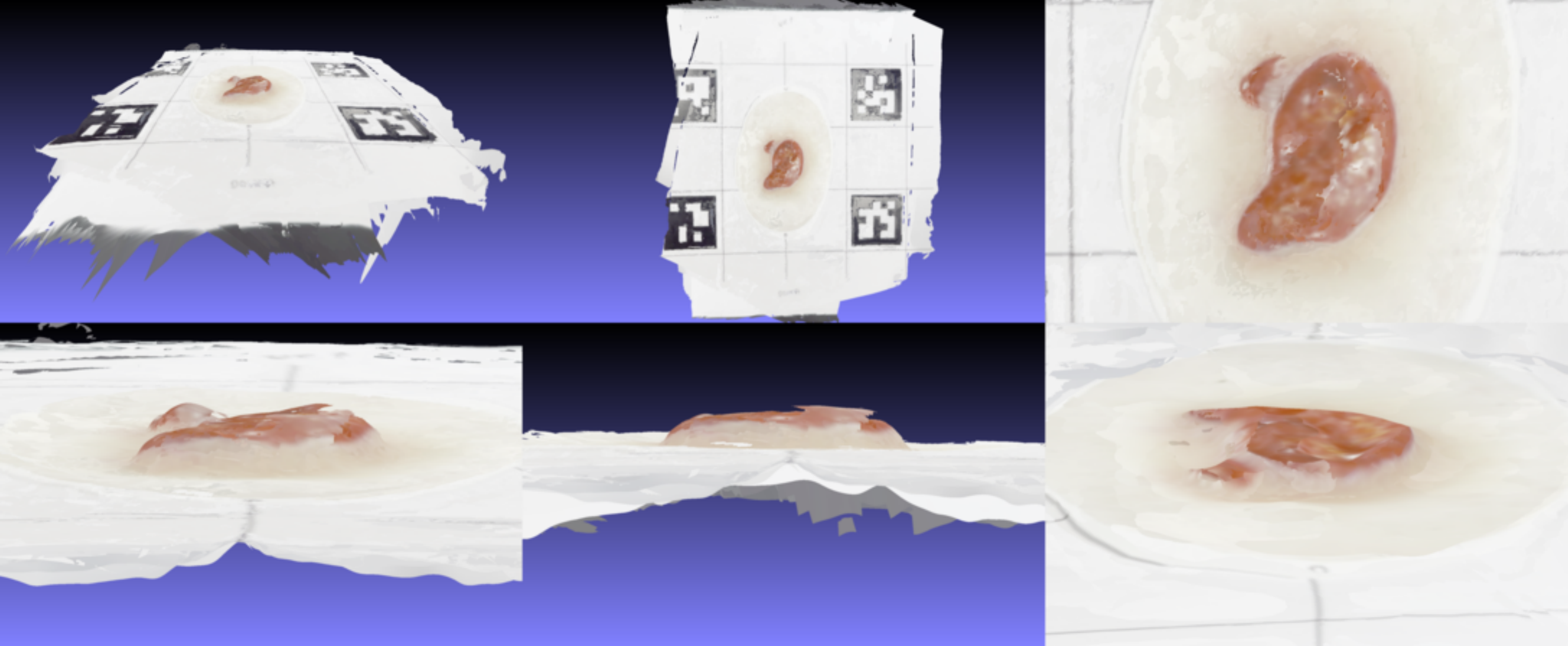}
    \caption{Six views of the mesh generated by DUSt3R for the PIS3 wound type. 
    From the middle figure of the second row, we can observe how DUSt3R does not retrieve accurate geometry. The perimeter of the wound is much higher than the surrounding regions, not reflecting the real geometry of the scene.
    DUSt3R, unlike traditional multi-view stereo approaches, do not follow epipolar constraints to generate the 3D structure but relies on transformers tailored to solve reconstructions from a few views in the wild.}
    \label{fig:duster}
\end{figure*}

\begin{figure*}[t!]
    \centering
    \includegraphics[width=\linewidth]{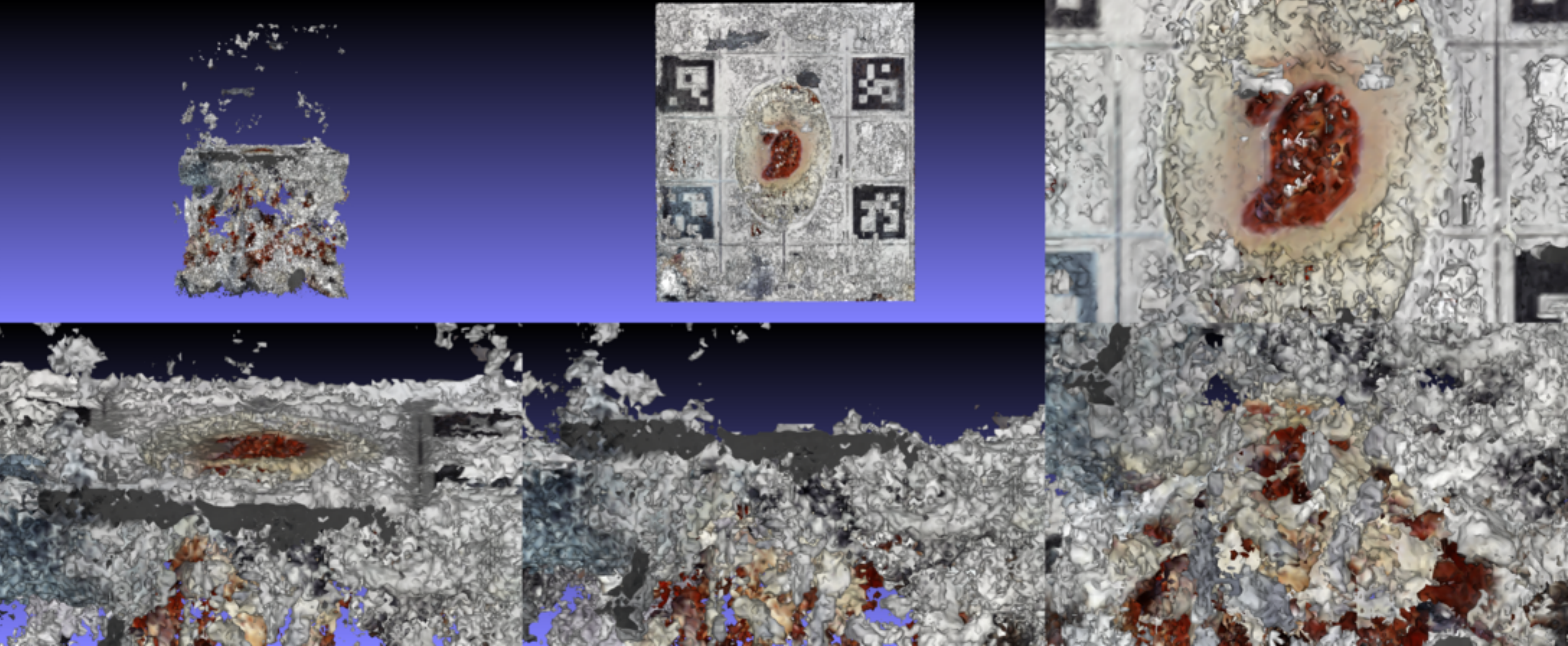}
    \caption{Six views of the mesh generated by Instant-NGP for the PIS3 wound type. As Instant-NGP is not a method developed for 3D surface reconstruction, its NeRF density is not regularised. As a result, when extracting a mesh using the Marching Cubes algorithm, it presents a structure similar to the one reported in the figure. Given the absence of surface regularizer terms, Instant-NGP solves photogrammetric consistency by placing density values under the wound's real surface, creating a scattered reconstruction.}
    \label{fig:ingp}
\end{figure*}

\begin{figure*}[t!]
    \centering
    \includegraphics[width=\linewidth]{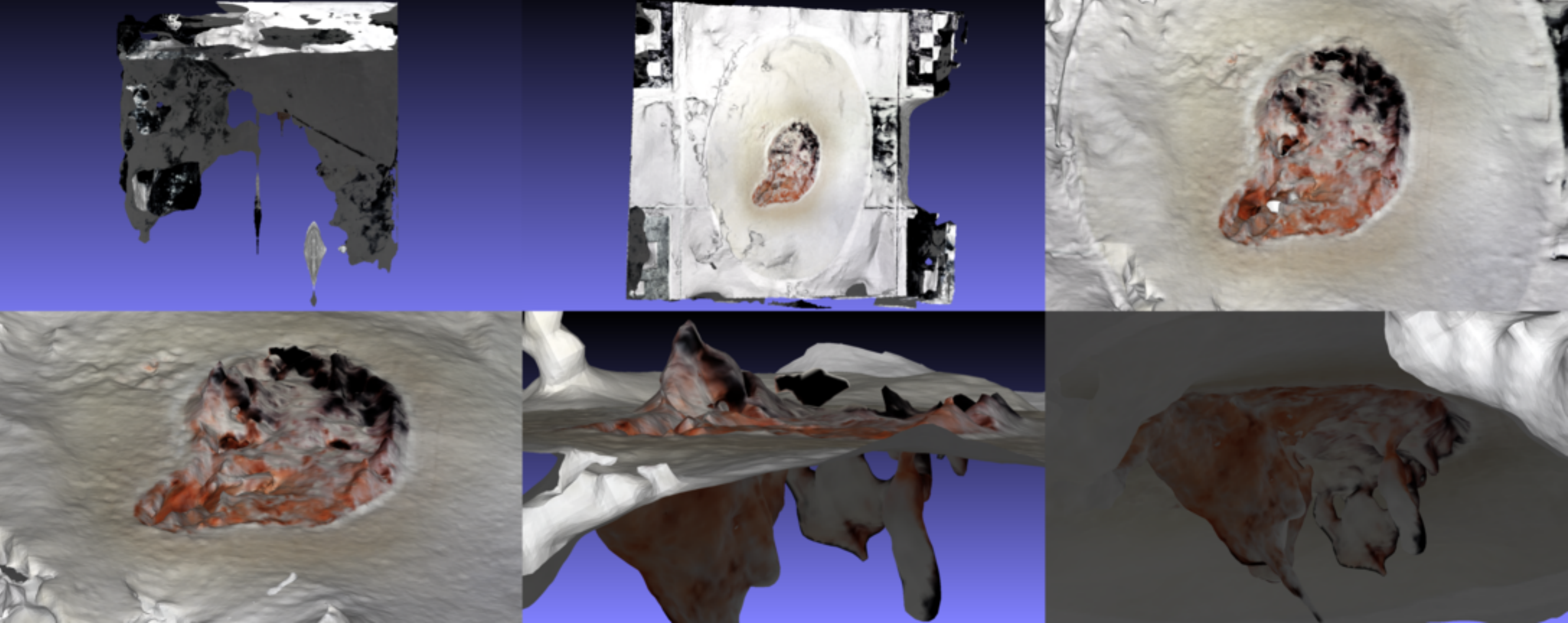}
    \caption{Six views of the mesh generated by NeuS2 for the PIS3 wound type. NeuS2 is a fast method for 3D surface reconstruction, however, we excluded it from our benchmark because it did not perform consistently well in our \NAME~dataset. Although more fine-tuning might be required for \NAME, we considered Neuralangelo and Neus-facto as better representing SDF methods in terms of robustness against photogrammetric complexities. For example, PIS3 presents hard reflections, as analysed in the renderings evaluation above, and we can observe in both figures on the right how NeuS2 attempts to satisfy photometric consistency and surface regularization by ``pushing'' the surface under the wound level. This behaviour can be attributed to its architecture similar to Instant-NGP.}
    \label{fig:nesu2}
\end{figure*}

\begin{figure*}[t!]
    \centering
    \includegraphics[width=\linewidth]{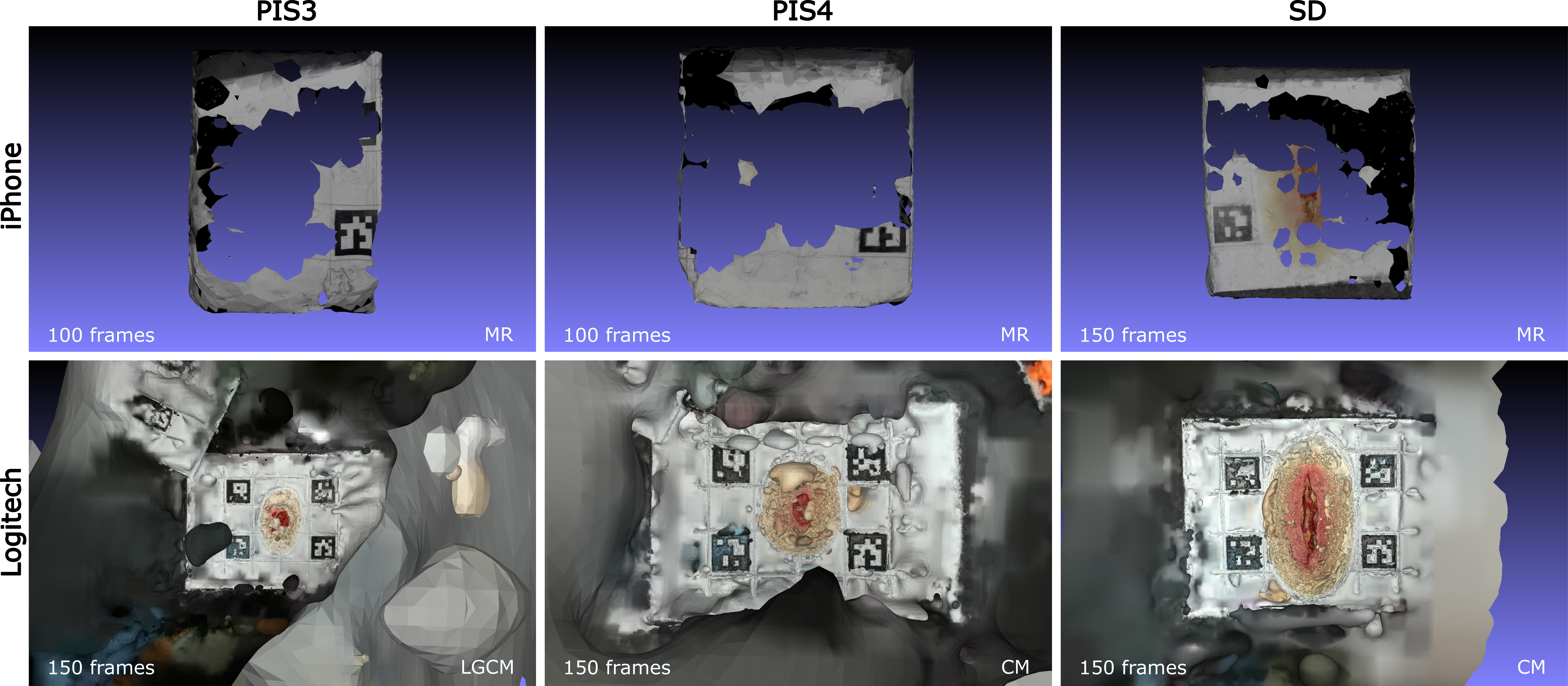}
    \caption{Examples of failed reconstruction while exploring how larger samples of images impact photogrammetric methods.
    In the first row, we display examples of 100 and 150 frames sampled from the iPhone sequence of each wound type. Meshroom (MR) was not able to reconstruct any detail in the wound and surrounding regions.
    In the second row, we display both COLMAP (CM) and COLMAP equipped with LightGlue feature matching (LGCM) at 150 frames. While both methods manage to reconstruct the wound area, the quality of the estimation drastically decreases compared to samples of 50 images.}
    \label{fig:higher_frames}
\end{figure*}

